\documentclass[12pt,a4paper]{article}
\usepackage[latin1]{inputenc}
\usepackage[T1]{fontenc}
\usepackage{pgf}
\usepackage{url}
\usepackage[english]{babel}
\usepackage{multirow}
\usepackage{fancyhdr}
\usepackage{anysize}
\usepackage{amssymb}
\usepackage{graphicx}
\usepackage[latin1]{inputenc}
\usepackage{url}
\usepackage{algorithm}
\usepackage{algpseudocode}
\usepackage[sort, numbers]{natbib}
\usepackage{ntheorem}

\begin{document}

\title{\bf{An eigenvector-based hotspot detection}}
\author{Hadi Fanaee-T and João Gama}
\date{}
\maketitle
\begin{center}
Laboratory of Artificial Intelligence and Decision Support (LIAAD), University of Porto\\
INESC TEC, Rua Dr. Roberto Frias, Porto, Portugal\\
hadi.fanaee@fe.up.pt and jgama@fep.up.pt\\
\end{center}

\begin{abstract}

Space and time are two critical components of many real world systems. For this reason, analysis of anomalies in spatiotemporal data has been a great of interest. In this work, application of tensor decomposition and eigenspace techniques on spatiotemporal hotspot detection is investigated. An algorithm called SST-Hotspot is proposed which accounts for spatiotemporal variations in data and detect hotspots using matching of eigenvector elements of two cases and population tensors. The experimental results reveal the interesting application of tensor decomposition and eigenvector-based techniques in hotspot analysis.
\end{abstract}


\section{Introduction}\label{tab:introduction}
Space and time are two critical components of many real world systems. Nowadays, there is great interest in spatiotemporal data analysis due to the huge amount of data available. Among various analysis tasks that can be carried out on spatiotemporal data, hotspot analysis is recognized as an important tool in security informatics and bio-surveillance. For instance in crime hotspot application, an outcome such as \textit{City Center between hours 8 to 11pm}  would be a spatiotemporal hotspot. Outcome like \textit{City Center} or \textit{City Park} would be strict spatial hotspots and \textit{8 to 11pm} and \textit{10 to 12pm} are samples of temporal hotspots.
Hotspot analysis goal consists of detecting spatiotemporal regions among data which are showing significant deviations comparing to the rest of the data.

In this paper we propose a novel spatiotemporal hotspot analysis approach based on tensor decomposition \cite{Kolda2008} and eigenvector elements matching. This approach can be applied for solving two types of problems. Firstly, it independently can detect semi-spatiotemporal hotspots and in the second application can be used as a pre-processing engine for common spatiotemporal methods to reduce a big search space to a few limited regions for post-processing. 

The rest of the paper is organized as follow. First in section \ref{sec:related} we outline the related works. In section  \ref{sec:algorithm} we explain the proposed algorithm for hotspot detection. In section \ref{sec:experiments} we describe the experiments and finally in section \ref{sec:conclusion} we end with conclusion and future works.

\section{Related Work} \label{sec:related}

Related spatiotemporal techniques can be divided into two main categories: scan statistics and clustering-based techniques. Clustering-based approaches \cite{Levine2006,Zeng2004,Jain1999,Birant20062,Kisilevich2006} are based on this idea that first, thresholds are inferred from the population data and then estimated thresholds are applied on clustering of data points of cases data. Clustering-based approaches have their own limitations and strengths. Their prominent benefit is that they provide exact shape of clusters opposed to the scan statistics-based methods where clusters necessarily should be a regular shape and are not realistic. On the other hand, handling complex data is not straightforward for clustering-based techniques.

\begin{table}[ht]
\begin{center}
\caption{Comparison between our approach and ST-Scan}\label{tab:stscancomparison}
		\begin{tabular}{ | p{5cm} | p{2.5cm} | p{3cm} | }
    \hline
    {\bf } & {\bf ST-Scan}  & {\bf Our approach}  \\ \hline
    Data input & Spatiotemporal  & Multidimensional \\ \hline
    Support additional attributes & No  & Yes \\ \hline
    Support additional dimensions & No  & Yes \\ \hline
    Spatiotemporal clusters & Exact & Approximate \\ \hline
    Automatic trend adjustment & No  & Yes \\ \hline
    Take into account interdependencies and correlation between dimensions and measurements & No  & Yes \\ \hline
    Automatically Noise reduction & No  & Yes \\ \hline
    \end{tabular}
\end{center}
\end{table}

Scan statistics  approaches are based on Kulldorff original work \cite{Kulldorff1998} and its variant extensions and are developed for detection of anomalous patterns from spatiotemporal data sets. Table \ref{tab:stscancomparison} demonstrates a comparison between our approach and Space-time scan statistics (ST-Scan) based approaches. As it can be seen, our approach not only have no restriction for the number of attributes and dimensions but also can discover interdependencies and correlations among different attributes and dimensions. This strength originates from tensor decomposition emphasis which already has shown a great performance in many real-world applications such as chemometrics, econometrics and psychometrics \cite{Johnson2007}. A comprehensive survey about tensor decomposition techniques and its applications is presented in \cite{Kolda2008}.
Tensor decomposition itself cannot be used for hotspot detection. We adapted some ideas from computer vision such as eigenvector-based model matching which are also shown promising results for face reorganization \cite{Park2000,Turk19911,Turk19912}. 
Another privilege of our approach comparing to the ST-Scan is its ability for reduction of the noises and taking into account the temporal adjustment in an automatic way. 

Our approach has some distinct features that make it distinguished from similar works. One of these distinctive features is its strength in handling interdependencies between attributes and dimensions in multilinear and multidimensional data. To the best of our knowledge, such feature is not taken into account in the existing approaches. This strength originates from tensor decomposition emphasis which already has shown a great performance in many real-world applications in particular areas such as chemometrics, econometrics and psychometrics \cite{Johnson2007}. However, tensor decomposition itself cannot be used for hotspot detection. We adapt some other ideas from computer vision such as eigenvector-based model matching which are also shown promising results for face reorganization \cite{Park2000,Turk19911,Turk19912}. Our work however is different from common eigenvector-based model matching in computer vision. We here, instead of comparing two eigenvectors together, compare their individual eigenvector elements.

\subsection{Space-time Scan Statistics (ST-Scan) }

There are several variants of ST-Scan approaches. Introducing and comparing of the work with all these dozens of variants is out of scope of this paper. So we confine ourselves to introduce the baseline ST-Scan method \cite{Kulldorff1998}. ST-Scan exhaustively moves a varying radius and height cylinder over spatiotemporal space where the height of this cylinder is corresponding to the time dimension and the surface covers the space dimension. It then computes a score F(S) based on the following likelihood ratio statistic \cite{Neill2006} (equation \ref{eq:stscan}) for each spatiotemporal cylinder:

\begin{equation} \label{eq:stscan}
F(S)=(\frac{C}{B})^{C}  (\frac{C_{total}-C}{B_{total}-B})^{C_{total}-C}
\end{equation}
\noindent where, \textit{C} is total counts and \textit{B} is baseline in \textit{S} and $C_{total}$ and $B_{total}$ are also total counts and baseline of search area respectively. Then all possible cylinders are sorted based on the highest to lowest score and a randomization test is performed for obtaining the cylinder statistical significance. Then spatiotemporal regions whose p-value is lower than a threshold (usually 0.05) are returned as spatiotemporal hotspots.

\section{Detection of Spatial and Temporal hotspots} \label{sec:algorithm}

\begin{algorithm}
\caption{SST-Hotspot} \label{algorithm:sst}
\begin{algorithmic}[1]

\Require Tensor P, Tensor C, Matrix Neighbors, Parameters [R1,R2, ... Rn] 
\Ensure LikelyCluster, FirstPriority, SecondPrority, TFirstPriority, TSecondPriority
\State Decompose P: $ES_p \gets 1st\ SEigenvector$, $ET_p \gets 1st\ TEigenvector$ \label{line:decompos1}
\State Decompose C: $ES_c \gets 1st\ SEigenvector$,  $ET_c \gets 1st\ TEigenvector$ \label{line:decompos2}
\State Perform Sign correction on Pair (ESp,ESc) and Pair (ETp,ETc) \label{line:sign}
\For{each region s}  \Comment{Spatial Eigenvector elements matching } \label{line:spatialmatch1}
	\State ${DS(s)} \gets ESp(s)-ESc(s)$
\EndFor \label{line:spatialmatch2}
\For{each time t} \Comment{Temporal Eigenvector elements matching } \label{line:temporalmatch1}
	\State ${DT(t)} \gets ETp(t)-ETc(t)$
\EndFor \label{line:temporalmatch2}
\State Descending Sort of DS  \label{line:s-same1} \label{line:spatial1} \Comment{Spatial}
\State ${SL} \gets DS(s)\succ0$ \label{line:SL}
\State $SC \gets Regions\  that\  have\ DS(s)>std(DS)$ \label{line:SC}
\State $ST \gets Remove\ SC\ from\ SL $ \label{line:ST}
\State ${S1} \gets Regions\ in\ ST\ that \ have\ DS(s) \geq std(ST)$ \label{line:S1}  \label{line:s-same2}
\State ${S2} \gets Regions\ in\ ST\ that\ have\ DS(s) \prec std(ST)$ \label{line:S2}
\State ${LikelyCluster} \gets Regions\ in\ S1\ with\ pair\ distance\ lower\ than\ std(ST)$ \label{line:MostLikely}
\For{each X in S1} \label{line:S1start}
		\For{each H in SC} 
					\State If Neighbors(X,H)==1 Then Add X to FirstPriority(H) 
		\EndFor
\EndFor		\label{line:S1end}
\State SecondPrority=FirstPriority \label{line:S2start}
\For{each X in S2} 
		\For{each H in SC} 
			\For{each M in FirstPriority(H)} 
					\State If Neighbors(X,M)==1 Then Add X to SecondPrority(H) 
			\EndFor
		\EndFor
\EndFor		\label{line:spatial2} \label{line:S2end}
\State Descending Sort of DT  \label{line:t-same1} \label{line:temporal1} \Comment{Temporal}
\State ${TL} \gets DT(t)\succ0$
\State $TC \gets Times\  that\  have\ DT(t)>std(DT)$
\State $TT \gets Remove\ TC\ from\ TL $
\State ${T1} \gets Times\ in\ TT\ that \ have\ DT(t) \geq std(TT)$ \label{line:t-same2}
\For{each H1 in TC} \label{line:T1start}
	\For{each H2 in TC} 
		\State Add non-repeated [H1-H2] to TFirstPriority
	\EndFor
\EndFor \label{line:T1end}
\For{each X in T1} \label{line:T2start}
	\For{each H in TC} 
		\State Add [X-H] to TSecondPriority
	\EndFor
\EndFor \label{line:temporal2} \label{line:T2end}

\end{algorithmic}
\end{algorithm}

\begin{figure}
 \begin{center}
  \includegraphics[width=9cm]{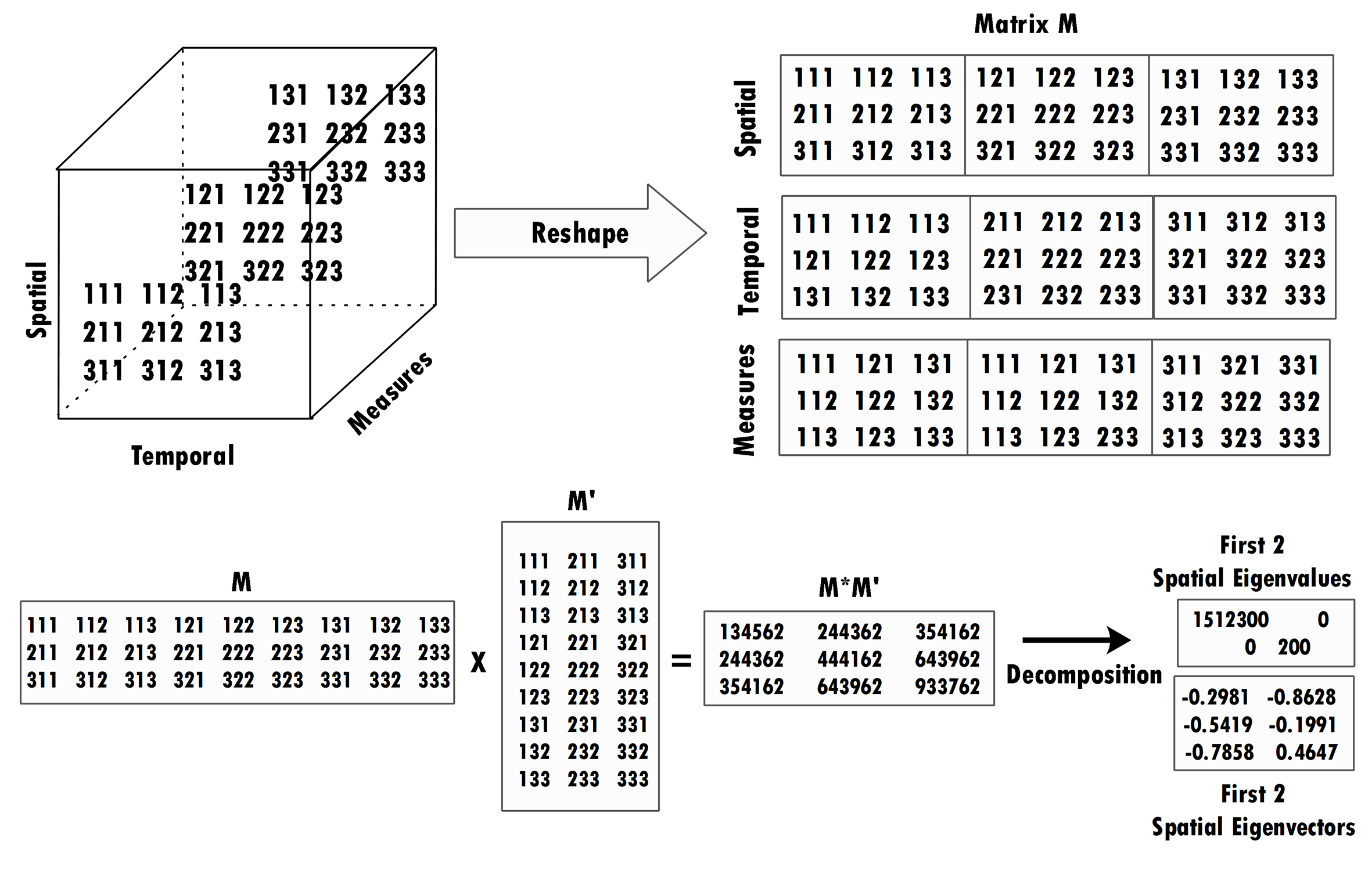}
 \end{center}
 \caption{A sample of 3D spatiotemporal tensor decomposition process} \label{fig:decomposition}
\end{figure}

Here we introduce our proposed algorithm for detection of spatial and temporal hotspots. As well as ST-Scan, our algorithm receives three data sets: population data, cases data and geographic data. Outcomes of the algorithm are spatial likely cluster, spatial first priority and second priority hotspot clusters and temporal hotspot intervals. The algorithm is composed of three main parts. The first part is tensor decomposition (lines \ref{line:decompos1}-\ref{line:decompos2}). The second part is eigenvector elements matching in lines \ref{line:spatialmatch1}-\ref{line:spatialmatch2} (spatial) and lines \ref{line:temporalmatch1}-\ref{line:temporalmatch2} (temporal) and the rest of the algorithm are allocated to spatial (lines \ref{line:spatial1}-\ref{line:spatial2}) and temporal (lines \ref{line:temporal1}-\ref{line:temporal2}) hotspot detection.

The required geographic data is a bit different from ST-Scan. In ST-Scan, geographic data includes coordinate of each region. This data need to be processed to be transformed to a neighbor's matrix. As a result we should have a matrix of regions-regions which each cell represents the boolean value such that 1 indicates that the region X is neighbor (e.g. has a border) with region Y.

Algorithm \ref{algorithm:sst} demonstrates the algorithm SST-Hotspot. As an input it receives population tensor, cases tensor, geographic data and model parameters. Model parameters \textit{Rn} by default for each dimension of space and time are 2 and for other dimensions are considered as 1 (R1=2, R2=2, R3=1, Rn=1, ... ). It means that tensor decomposition will output 2 eigenvectors for space and time and one for other dimensions. 

At lines \ref{line:decompos1}-\ref{line:decompos2} of algorithm we decompose population and cases tensor. \textit{SEigenvector} and \textsl{TEigenvector} represents spatial and temporal eigenvectors respectively. This process is illustrated in Figure \ref{fig:decomposition}. A 3D spatiotemporal tensor with dimensions of $3\times3\times3$ (spatial, temporal and measures) is demonstrated in this figure. We first reshape population and cases tensor in each dimension to the 2D matrix M. Thereafter, we first form a matrix of M$\times$M' and then perform a matrix decomposition to obtain its first 2 eigenvectors. The reason why we retrieve only 2 eigenvectors is that our model parameter for spatial mode has been set to 2 in this example. The First eigenvector represents majority of variance in data and is appropriate for model comparison. Second and rest eigenvectors only include small variations of data and might include noises. We do not limit the algorithm to only one eigenvector. However based on our preliminary observations, on the case study data set, taking the first eigenvector was appropriate for model matching. In other cases one might need to pre-examinate the model on a sample data to see how many eigenvectors are appropriate for model matching.

\begin{figure}
 \begin{center}
  \includegraphics[width=11cm]{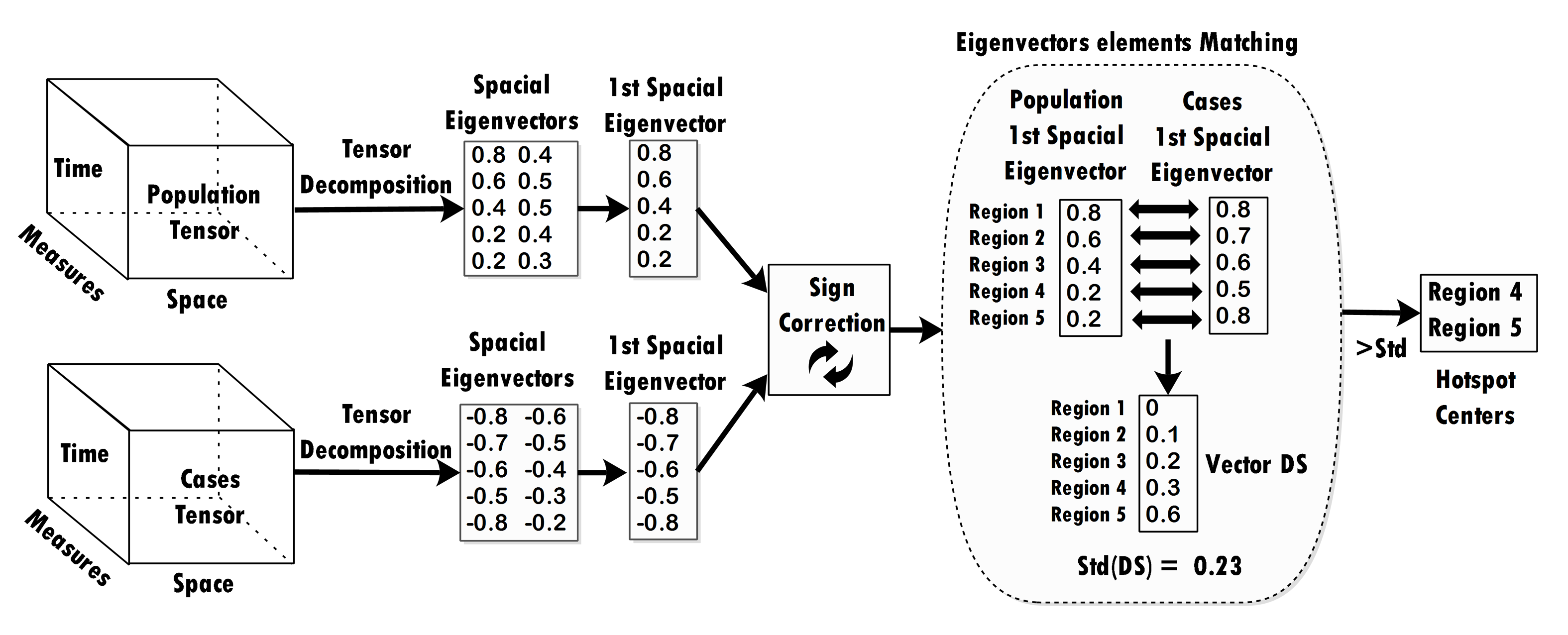}
 \end{center}
 \caption{A sample of Eigenvectors elements matching process} \label{fig:hotspotcenters}
\end{figure}

After decomposition step we need to perform a sign correction \cite{Park2000,Shapiro2000} at line \ref{line:sign}. The reason why we need to apply sign correction is that eigenvectors are not unique; i.e., if \textit{ei} is an eigenvector, then -\textit{ei} also can be an eigenvector. Therefore, without sign correction or alignment of the corresponding dimension for the two models, a direct comparison of eigenvectors is meaningless \cite{Park2000}. We adapt sign correction technique from \cite{Park2000} to make both sets of eigenvectors have consistent directions. 

At lines \ref{line:spatialmatch1}-\ref{line:temporalmatch2} for each spatial region and each temporal point we perform  eigenvector elements matching. Note that we do not compute the distance between two eigenvectors rather we match each eigenvector elements corresponding to each region. This process is illustrated in Figure \ref{fig:hotspotcenters}. This figure shows two same size tensors of population and cases. As is illustrated, in order to obtain first two eigenvectors, population and cases tensors are decomposed and after sign correction, value of each region eigenvector element in cases first eigenvector is subtracted from the corresponding one in population first eigenvector. This is repeated for all regions and obtained values are allocated in vector DS. Then, the standard deviation of this vector is computed as a threshold. Those regions that their corresponding distances are greater than this threshold are reported as hotpots centers (e.g. in Figure \ref{fig:hotspotcenters}, region 4 and 5) which is equivalent to SC in the algorithm. 

At line \ref{line:SL} we make another copy of DS saving in variable SL where DS(s) is greater than zero. At line \ref{line:SC} we identify hotspot centers as those regions which DS(s) are greater than standard deviation of DS. At line \ref{line:ST} we make another copy of SL by excluding of hotspot centers into variable ST. We then define standard deviation of ST (std(ST)) as a threshold that distinguish first priority (S1) from second priority (S2) list. Based on this, S1 would include regions whose DS(s) are greater than std(ST) (line \ref{line:S1}) and S1 includes regions whose DS(s) are lower than std(ST) (line \ref{line:S2}). We define likely cluster at line \ref{line:MostLikely} as hotspots which are shown up in the S1 and their pair distances are very close (lower than std(ST)).

In order to detect hotspot clusters, in the first step (lines \ref{line:S1start}-\ref{line:S1end}) we connect regions that are appeared in S1 to one of hotspot centers if they are spatial neighborhood. In other words, if for instance region A is identified as a hotspot center and its spatial neighbor B is appeared in S1, likely they form a hotspot cluster FirstPriority(1)=\{A,B\}. Although, the first priority list includes the most abnormal regions, however those regions that their DS(s) is greater than zero would be interesting for some purposes. For this reason we generate another list called second priority hotspot clusters. We first make a copy of first priority clusters and then connect those regions that are appeared in S2 to one of first priority cluster members. This is done via lines \ref{line:S2start}-\ref{line:S2end} in the algorithm. The process is the same as first priority list with this difference that this time we consider the new members of first priorities clusters. For instance in the above example, suppose that region C is not appeared in the S1 but is appeared in S2 and spatially is neighborhood of region B. So second priority list would be SecondPriority(1)=\{A,B,C\}. Now suppose that we have another region D that is appeared in S2 and is a neighborhood of region C but none of regions A and B. This region is not added to second priority list, because it has no connection with the first priority members A and B.

Detection of temporal hotspots is similar to the spatial one. Explanation of lines \ref{line:t-same1}-\ref{line:t-same2} are almost same as lines \ref{line:s-same1}-\ref{line:s-same2}. The only difference is this that for temporal detection we do not look for clusters; rather we only need to connect temporal hotspots together to generate hotspot intervals (e.g. '86-89'). At line \ref{line:T1start}-\ref{line:T1end} we generate intervals between each temporal hotspot centers together and in lines \ref{line:T2start}-\ref{line:T2end} we make an interval between each point in T1 and one of temporal hotspot centers. 

\section{Experiments} \label{sec:experiments}

We applied our algorithm on a real data set and tried to compare its performance in comparison to ST-Scan as a basline method to assess the algorithm performance. We used MATLAB running on a personal PC with Intel Core 2 Duo CPU and 3GB Ram. Three MATLAB toolboxes were also used during the experiments: Tensor toolbox \cite{Bader2012}, ITA toolbox \cite{Sun2012} and mapping toolbox \cite{Mapping2012} for drawing the results on the map. In the following we introduce the used data set and obtained results.

\subsection{Data Set}

We took the Incidence and Population data from New Mexico brain cancer data set used in \cite{Kulldorff1998}. The original source of data is \textit{Surveillance, Epidemiology and End Results (SEER) program} of national cancer institute, collected by New Mexico tumor registry between years of 1973 to 1991 for 32 sub-regions of New Mexico State, United states. There are 1175 reported cases of malignant neoplasm of the brain and nervous system. Each record of cases data includes region of residence, year of diagnosis, age group in 5-year interval (19 groups), race (white, black, other) and sex. Similarly the same attributes are available for populations. Data set is publicly available at \cite{Kulldorff2012}. We also used PostgreSQL spatial extension \cite{PostGIS2012} on New Mexico state shape file to to extract the neighborhood matrix. After all we transformed data sets to the tensor format. We made three tensors in different orders: 2D \{D1=Space, D2=Time\} where we ignored other attributes such as age, sex and race, 3D tensor \{D1=Space, D2=Time, D3= (Age, Sex, Race)\} and 5D \{D1=Space, D2=Time, D3=Age, D4=Sex, D5=Race\}. Each cell in above tensors represents the count (cancer observations in case set and population count in population set).

\subsection{Results} \label{sec:results}

Experiments were designed to answer the following questions. In the following subsections we discuss the obtained results regarding each research question.

\begin{enumerate}
	\item What is the benefit of higher order data modeling? How we can choose the order?
	\item What is the validity of the detected Hotspots?
	\item How does our approach prune non-significant regions?
	\item How does our approach treat with temporal trends in population data?
\end{enumerate}

\begin{figure}[ht]
 \begin{center}
  \includegraphics[width=11cm]{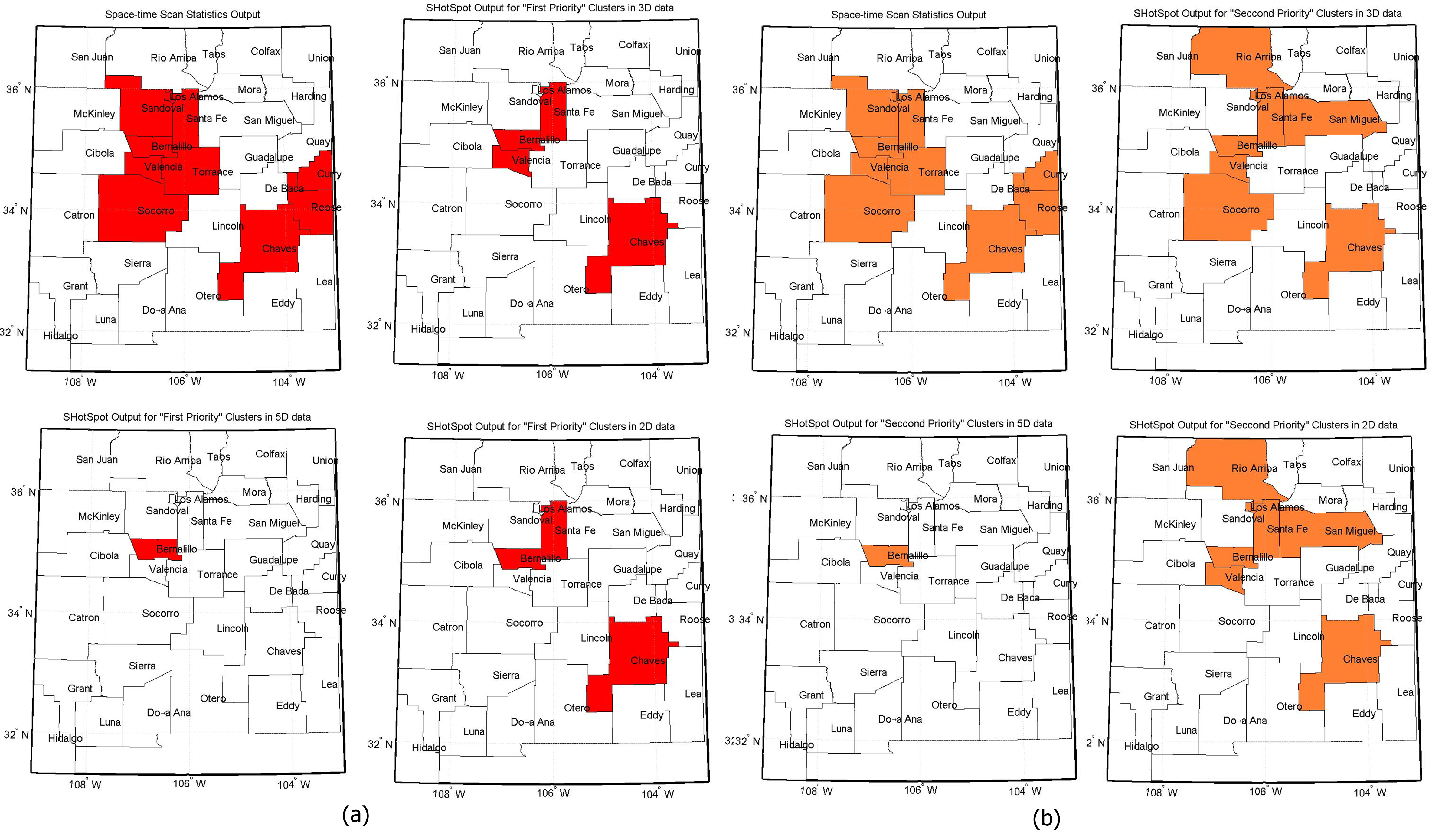}
 \end{center}
 \caption{Hotspot clusters, ST-Scan (Top Left) , SST-Hotspot 2D (Bottom Right), SST-Hotspot 3D (Top Right) and SST-Hotspot 5D (Bottom Left). (a) First priority clusters and (b) Second priority clusters } \label{fig:map}
\end{figure}

We applied our algorithm on the generated 2D, 3D and 5D tensors and then compared the first and second priority hotspot clusters to the ST-Scan outputs presented in \cite{Kulldorff1998} (which is already obtained via SatScan software). Results for first priority clusters are presented in Figure \ref{fig:map}.a and for second priority cluster are presented Figure \ref{fig:map}.b. We used standard metrics such as F1 (F-measure), precision and recall as well for comparing output of the algorithm with ST-Scan. This result is shown in Table \ref{tab:compare1}. Likely detected cluster from ST-Scan and SST-Hospot on different tensor orders are also reported in Table \ref{tab:detected}. 

\subsubsection{Validation of hotspots by comparing with ST-Scan results.}

Looking to the Table \ref{tab:detected} we easily can observe that how our algorithm detected the important spatial regions "\textit{Bernalillo} and \textit{Chaves}" and two important temporal hotspot centers: \textit{85} and \textit{89}. As it can be seen, this result is similar to ST-Scan. ST-Scan outputs another three clusters "\textit{Chaves}" , "\textit{Curry and Roosevelt}" and "\textit{Los Alamos}". Among these clusters, SST-Hotspot successfully detected "\textit{Chaves}". However, "\textit{Curry} and \textit{Roosevelt}" are neighbors of \textit{Chaves} and \textit{Los Alamos} is almost near to the \textit{Bernalillo}. In other words, SST-Hotspot effectively detected the center of events both spatially and temporally. If we look to the obtained result from ST-Scan (Top left figure in Figures \ref{fig:map}.a and  Figure \ref{fig:map}.b) we find two major affected zones with center of \textit{Bernalillo} and \textit{Chaves}. Both of these centers are detected by SST-Hotspot. As was expected, the most significant anomalous parts of data are appeared in the eigenspace.

\begin{table}[ht]
\centering
\caption{Comparisons of detection power of spatial hotspots with ST-scan results in each tensor order} 
		\begin{tabular}{|p{2.5cm}|p{1.5cm}|p{1.5cm}|p{1.5cm}|p{1.5cm}|p{1.5cm}|p{1.5cm}|}
        \hline
	       ~ &  \multicolumn{2}{|c|}{SST-Hotspot 2D} & \multicolumn{2}{|c|}{SST-Hotspot 3D}& \multicolumn{2}{|c|}{SST-Hotspot 5D} \\ \hline
         ~ & First Priority & Second Priority & First Priority  &  Second Priority  & First Priority & Second Priority \\ \hline
        Precision & 100.0       & 71.43        & \textbf{100.00} & \textbf{75.00} & 100.00       & 100.00  \\ \hline
        Recall    & 30.00        & 50.00        & \textbf{40.00}  & \textbf{60.00} & 10.00        & 10.00   \\ \hline
        F1        & 46.15        & 58.82        & \textbf{57.14} & \textbf{66.67} & 18.18        & 18.18   \\ \hline
    \end{tabular}
\label{tab:compare1}
\end{table}

\begin{table}
\centering
\caption{Detected Hotspot Clusters}
\begin{tabular}{|p{3cm}|p{2.5cm}|p{2cm}|p{2cm}|p{2cm}|}
\hline
Method &  ST-Scan & SST-Hotspot 2D & SST-Hotspot 3D & SST-Hotspot 5D \\ \hline
Likely reported spatial clusters & Bernalillo (center of cluster) & Bernalillo & Bernalillo & Bernalillo \\ \hline
Secondary reported spatial clusters & Chaves , Curry and Roosevelt , Los Alamos & Chaves & Chaves & Not Found \\ \hline
Likely reported spatial cluster with trend adjustment & Los Alamos and SantaFe  & Not Found & Los Alamos and SantaFe & Not Found \\ \hline
Likely reported temporal clusters & 85-89 & 85,89 & 85,89 & 86,89 \\ \hline
Secondary reported temporal clusters & 88-89 , 82-83 , 86-89 & 86,90,88,83 & 86,88,83,90 & 88,85,91 \\ \hline
\end{tabular}
\label{tab:detected}
\end{table}

\subsubsection{Effects of different tensor ordering.}

ST-Scan by default does not consider the existing trend in population data. The solution provided in \cite{Kulldorff1998} was to increase incidence rates 1.2 percent per year based on the available growth rate in population data. Based on this adjustment the only reported cluster from SatScan Software is "\textit{Los Alamos} and \textit{SantaFe}" in years "\textit{86-89}". Our algorithm SST-Hotspot on 3D order also could detect this hotspot. For instance in our experiments on 3D ordering and in S1 list we obtained DS(Los Alamos)=0.005 and DS(Santa Fe)=0.004 where standard deviation of DS vector was equal to Std(DS)=0.004. There were only these two regions in S1 list that their distances together was lower than standard deviation of DS vector (|0.004-0.005|=0.001<0.004) and therefore were reported as most likely hotspot at line \ref{line:MostLikely} of algorithm. This is while in other orders 2D and 5D, we could not detect this hotspot. This shows the importance of tensor order selection. 

The reason why this cluster is not detected in 2D and 5D is this that ST-Scan adjusts counts for sex, age and race and thus naturally effects of these attributes are considered in the counts, so, as was expected, the results had to become more similar to ST-Scan. 2D ordering do not take into account other attributes (sex, age and race) and clearly shall not detect similar hotspots as well as ST-Scan. In terms of 5D ordering, the fit of the model for cases data was very low (about 14 percent). For this reason model obtained from cases tensor would not be a good sample to compare with population model with fit of over 89 percent. The reason is this fact that decomposition methods performance drops when face with a sparse tensors. This is while we get 61 percent for cases tensor and 96 percent for population tensor in 3D order. Based on the above findings we conclude that higher order modeling of data improves the detection power, however the ordering needs to be selected carefully. For instance in our case study, the highest tensor order was 5 but presented a lower quality than order 3.

\subsubsection{Pruning functionality evaluation.}

As we earlier mentioned, our algorithm has two functions, first it can be used as a method for spatial and temporal hotspot detection from multidimensional data and second can be used as a pre-processing step in ST-scan process to reduce the search space. It can prune thousands of non-significant spatial or temporal regions in ST-Scan. This means that more cover is required and thus the recall measure would be more significant for us. Because when it is used in ST-Scan it should cover more candidate regions for being tested in ST-Scan process. Therefore the accuracy does not make scene because the final decision about anomaly of the regions should be made by ST-Scan test.

However when SST-Hotspot is going to be applied independently, F-measure need to be taken into account instead. The best F-measure obtained for first priority list of 3D model was 66.67 percent. It means that when the goal is to detect the spatial hotspots from spatiotemporal data we can expect a close result to SST-Scan. This is extremely reasonable when we compare the required extensive computation cost of ST-Scan with light computation cost of SST-Hotspot. 

As a result, instead of performing ST-scan test for thousands of cylinders we just need to run the test for a limited regions and temporal periods outputted from SST-Hotspot. In fact, we have the center of events and suspicious neighbors and suspicious time intervals in SST-Hotsput output. So, we only need to justify the cylinder size based on obtained suspicious places and temporal intervals. For instance, see Table \ref{tab:detected}. As it can be seen, the main center of most likely cluster (\textit{Bernalillo}) and the two sides of temporal interval of most likely temporal cluster (\textit{85} and \textit{89}) are detected through SST-Hotspot. So a cylinder with circle with center of \textit{Bernalillo} with radius from the most far cluster members (\textit{Los Alamos} and \textit{Socorro}) and height of 85 to 89 (4 years as a size of time window) can be a good primary guess for limiting ST-Scan search space. For other clusters for instance, \textit{Chaves and Roosevelt} and \textit{Curry} we just need to put the center of cylinder surface on the \textit{Chaves} county with each of obtained temporal hotspots from SST-Hotspot first priority list (86-89, 88-89, 88-89 and 83-89). In this case through only four tests we detect \textit{Chaves} cluster in year 88-89. 

\subsubsection{Trends handling ability.}

If we configure ST-Scan for temporal trend justification, only one cluster is detected: \textit{Los Alamos and Santa Fe in years \textit{86-89}}. ST-Scan in fact increases cases counts equal to 1.2 percent annually because of growing trend in population. For this reason its output is different from when this temporal trend justification is not performed. Our algorithm on 3D data could also detect this spatial hotspot \textit{Los Alamos and Santa Fe} at the same time. However this was not detected in 2D and 5D data. One obvious reason is that this cluster in ST-Scan results is obtained based on adjusted counts for age, sex and race. So, 2D data which ignore all other attributes, definitely shall not detect that. 5D data is also as previously mentioned would not be able to detect that because of high fit difference between its population and cases models.  The reason why 3D model could detect that is due this fact that it takes into account the correlation between all variables with time. During the decomposition process when we just select first eigenvector. So noises and other hidden factors are removed from the model. Thus, logically SST-Hostspot 3D shall detect clusters similar to ST-Scan.

\section{Conclusion and Future works} \label{sec:conclusion} 

In this paper we presented a novel approach for hotspot detection exploiting tensor decomposition and eigenvector elements matching techniques. The experimental results reveal the effectiveness of these techniques. Our approach is not a replacement for ST-Scan and its variants rather can be considered as a helpful method to reduce the search space in ST-Scan process, even though one could adapt this method independently for monitoring spatiotemporal variance in data. We also showed that how multi-way data analysis improves the quality of detection as it already was expected. One major of our approach drawback, is its inability about automatically connecting of the separated spatial and temporal hotspots with aim of identification of spatiotemporal hotspots. Another issue is that the current version is retrospective. Adapting algorithm for online detection would be another feature research direction.

\subsubsection*{Acknowledgments.} This work is funded by the European Regional Development Fund through the COMPETE Programme, by the Portuguese Funds through the FCT (Portuguese Foundation for Science and Technology) within project FCOMP - 01-0124-FEDER-022701.

\bibliographystyle{chicago}
\bibliography{ref}
\end{document}